\documentclass{article}
\usepackage{ijcai25}

\usepackage{times}
\usepackage{soul}
\usepackage{url}
\usepackage[hidelinks]{hyperref}
\usepackage[utf8]{inputenc}
\usepackage[small]{caption}
\usepackage{graphicx}
\usepackage{amsmath}
\usepackage{amsthm}
\usepackage{booktabs}
\usepackage{algorithm}
\usepackage{algorithmic}
\usepackage[switch]{lineno}
\usepackage[caption=false,font=footnotesize,labelfont=rm,textfont=rm]{sub
fig}
\usepackage{multirow} 
\usepackage{xcolor}
\usepackage{amssymb}
\usepackage[sectionbib]{chapterbib}

\title{GSDet: Gaussian Splatting for Oriented Object Detection}

\author{
Zeyu Ding$^{1,2}$\and
Jiaqi Zhao$^{1,2}$\thanks{Corresponding author}\and
Yong Zhou$^{1,2}$\and
Wen-liang Du$^{1,2}$\and
Hancheng Zhu$^{1,2}$\and
Rui Yao$^{1,2}$\\
\affiliations
$^1$School of Computer Science and Technology, China University of Mining and Techology\\
$^2$Mine Digitization Engineering Research Center of the Ministry of Education\\
\emails
\{dingzeyu, jiaqizhao, yzhou, zhuhancheng, wldu, ruiyao\}@cumt.edu.cn
}

\newcommand{\tablestyle}[2]{\setlength{\tabcolsep}{#1}\renewcommand{\arraystretch}{#2}\centering\footnotesize}

\usepackage{color, colortbl}

\newcolumntype{x}[1]{>{\centering\arraybackslash}p{#1pt}}
\newcolumntype{y}[1]{>{\raggedright\arraybackslash}p{#1pt}}
\newcolumntype{z}[1]{>{\raggedleft\arraybackslash}p{#1pt}}

\newcommand\dunderline[2][.4pt]{%
  \raisebox{-#1}{\underline{\raisebox{#1}{\smash{\underline{#2}}}}}}

\newcommand{\GSDettrain}{
    \begin{algorithm}[tb]
        \caption{GSDet Training}
        \label{alg:gsdet_train}
        \textbf{Input}: images, GT\\
        \textbf{Output}: loss $\mathcal{L}$
        \begin{algorithmic}[1] 
            \STATE $\{P_2,\dots, P_6\}$ $\leftarrow$ Encoder(images)
            \STATE $\{F_2,\dots, F_6\}$ $\leftarrow$ 3DFeatureSpace($\{P_2,\dots, P_6\}$)
            \STATE Initial3DGau $\leftarrow$ Random Initialization
            \STATE Result3DGau $\leftarrow$ Decoder($\{F_2,\dots, F_6\}$, Initial3DGau)
            \STATE Result2DGau $\leftarrow$ Splatting(Result3DGau)
            \STATE GT2DGau $\leftarrow$ Transform(GT)
            \STATE $\mathcal{L}$ $\leftarrow$ Loss(Result2DGau, GT2DGau)
            \STATE \textbf{return} $\mathcal{L}$
        \end{algorithmic}
    \end{algorithm}
}

\newcommand{\DIORRResults}{
    \begin{table*}[t]
    \centering
    \resizebox{0.99\linewidth}{!}
    {
    \begin{tabular}{ccccccccccccc}
    \toprule[1.02pt]
    \multirow{2}{*}{\textbf{Type}}&\multirow{2}{*}{\textbf{Method}} & \multirow{2}{*}{\textbf{Backbone}} & \multicolumn{4}{c}{\textbf{DIOR-R}} & \multicolumn{3}{c}{\textbf{DOTA-v1.0}} & \multicolumn{3}{c}{\textbf{DOTA-v1.5}}\\ \cmidrule[0.8pt](lr){4-7}\cmidrule[0.8pt](lr){8-10}\cmidrule[0.8pt](lr){11-13}
     & & & \textbf{AP$_{50}$} & \textbf{AP$_{75}$} & \textbf{mAP} & \textbf{FPS}&\textbf{AP$_{50}$} & \textbf{AP$_{75}$} & \textbf{mAP} &\textbf{AP$_{50}$} & \textbf{AP$_{75}$} & \textbf{mAP} \\ 
    \midrule[0.8pt]
    \textit{One-stage}&Oriented Rep~\cite{oriented_reppoints} & R50 & 66.31 & 44.36 & 42.81 & 29 & 74.38 & 46.56 & 44.57  & 64.04 & 38.05 & 37.43 \\
    \textit{One-stage}&DCFL~\cite{dcfl}  & R50 & 66.41 & 41.83 & 40.88 & \dunderline{29} & 73.13 & 39.62 & 41.14  & 67.00 & 39.13 & 38.60 \\
    \textit{One-stage}&H2RBox-v2~\cite{h2rboxv2} & R50 & 57.13 & 32.13 & 32.77 & \textbf{30} & 72.60 & 39.67 & 40.53  & 64.71 & 32.18 & 34.97 \\
    \textit{Two-stage}&COBB-sig~\cite{cobb} & R50 & 65.65 & 42.78 & - & 14 & \dunderline{75.52} & 48.35 & 45.61  & 66.25 & 41.34 & 40.04 \\
    \textit{Query}&ARS-DETR~\cite{arsdetr}  & R50 & 66.12 & 45.81 & 43.89 & 12 & 74.16 & 49.41 & 46.21  & 64.29 & 38.23 & 37.08 \\
    \textit{\textbf{GS}}&\textbf{GSDet  (900@1)} & R50 & \dunderline{69.35} & \dunderline{48.27} & \dunderline{46.19} & 28 &75.44 & \dunderline{52.15} & \dunderline{47.77}  & \dunderline{67.29} & \dunderline{42.49} & \dunderline{41.11}   \\
    \textit{\textbf{GS}}&\textbf{GSDet  (900@3)} & R50 & \textbf{70.05} & \textbf{48.91} & \textbf{46.69} & 21 & \textbf{75.74} & \textbf{52.40} & \textbf{48.16}  & \textbf{67.84} & \textbf{43.69} & \textbf{41.60}  \\
    \bottomrule[1.02pt]
    \end{tabular}
    }
    \caption{Comparisons with different oriented object detectors on \textbf{DIOR-R}, \textbf{DOTA-v1.0}, and \textbf{DOTA-v1.5}.}
    \vspace{-0.5em}
    \label{tab:diorr}
    \end{table*}
}

\newcommand{\DOTAOneResults}{
    \begin{table*}[t]   
    \centering
    \tablestyle{3pt}{1}
    \resizebox{1.0\textwidth}{!}
    {
        \begin{tabular}{cccccccccccccccccccc} 
        \toprule[1.1pt]
        & \textbf{Method} & \textbf{B.} & \textbf{PL} & \textbf{BD} & \textbf{BR} & \textbf{GTF}  & \textbf{SV} & \textbf{LV} & \textbf{SH} & \textbf{TC} & \textbf{BC} & \textbf{ST} & \textbf{SBF}   & \textbf{RA} & \textbf{HA} & \textbf{SP} & \textbf{HC} &\textbf{AP$_{50}$}\\ 

          \midrule[0.8pt]
          \multirow{7}{*}{\rotatebox{90}{\textbf{\textit{One-Stage}}}}&PSC~\cite{psc}      & R50   & 88.24 & 74.42 & 48.63 & 63.44 & 79.98 & 80.76 & 87.59 & 90.88 & 82.02 & 71.58 & 59.12 & 60.78 & 65.78 & 71.21 & 53.06 & 71.83 \\
          &R3Det~\cite{r3det}    & R101   & 88.76 & 83.09 & 50.91 & 67.27 & 76.23 & 80.39 & 86.72 & 90.78 & 84.68 & 83.24 & 61.98 & 61.35 & 66.91 & 70.63 & 53.94 & 73.79 \\
          &S2A-Net~\cite{s2anet}   & R50  & 89.11 & 82.84 & 48.37 & 71.11 & 78.11 & 78.39 & 87.25 & 90.83 & 84.90 & 85.64 & 60.36 & 62.60 & 65.26 & 69.13 & 57.94 & 74.12 \\
          &H2RBox~\cite{h2rbox} & R50   &88.93 & 78.89 & 46.27 & 68.79 & \dunderline{81.12} & 75.45 & 86.68 & 90.89 & 86.71 & \textbf{87.33} & 64.15 & \dunderline{68.83} & 62.81 & 69.39 & 59.79 & 74.40 \\
          &DHRec~\cite{dhrec}   & R50  & 88.58 & 77.90 & 53.84 & 72.93 & 78.45 & 78.84 & 87.64 & 90.88 & \textbf{88.78} & 85.46 & 56.11 & 66.74 & 67.58 & 70.25 & 57.53 & 74.57 \\
          &SASM~\cite{sasm}     & R50   & 86.42 & 78.97 & 52.47 & 69.84 & 77.30 & 75.99 & 86.72 & 90.89 & 82.63 & 85.66 & 60.13 & 68.25 & 73.98 & 72.22 & 62.37 & 74.92 \\
          &FRED~\cite{fred} & Re50 & 89.37 & 82.12 & 50.84 & 73.89 & 77.58 & 77.38 & 87.51 & 90.82 & 86.30 & 84.25 & 62.54 & 65.10 & 72.65 & 69.55 & 63.41 & 75.56 \\
        \midrule[0.8pt]
          \multirow{6}{*}{\rotatebox{90}{\textbf{\textit{Two-Stage}}}}  &SCRDet~\cite{scrdet}     & R101 & \textbf{89.98} & 80.65 & 52.09 & 68.36 & 68.36 & 60.32 & 72.41 & 90.85 & 87.94 & 86.86 & \dunderline{65.02} & 66.68 & 66.25 & 68.24 & 65.21 & 72.61 \\
          &RoI Trans.~\cite{roi_transformer}   & R50   & 88.65 & 82.60 & 52.53 & 70.87 & 77.93 & 76.67 & 86.87 & 90.71 & 83.83 & 52.81 & 53.95 & 67.61 & 74.67 & 68.75 & 61.03 & 74.61 \\
          &Gliding V.~\cite{gliding_vertex} & R101   & \dunderline{89.64} & \dunderline{85.00} & 52.26 & \dunderline{77.34} & 73.01 & 73.14 & 86.82 & 90.74 & 79.02 & 86.81 & 59.55 & \textbf{70.91} & 72.94 & 70.86 & 57.32 & 75.02 \\
          &AOPG~\cite{dior} & R50 & 89.27 & 83.49 & 52.50 & 69.97 & 73.51 & 82.31 & 87.95 & 90.89 & 87.64 & 84.71 & 60.01 & 66.12 & 74.19 & 68.30 & 57.80 & 75.24 \\
          &DODet~\cite{dodet}  & R50 & 89.34 & 84.31 & 51.39 & 71.04 & 79.04 & 82.86 & 88.15 & \textbf{90.90} & 86.88 & 84.91 & 62.69 & 67.63 & 75.47 & 72.22 & 45.54 & 75.49 \\
          &O. R-CNN~\cite{orientedrcnn}& R50   & 89.46 & 82.12 & \dunderline{54.78} & 70.86 & 78.93 & 83.00 & 88.20 & \dunderline{90.90} & 87.50 & 84.68 & 63.97 & 67.69 & 74.94 & 68.84 & 52.28 & 75.87 \\
        \midrule[0.8pt]
          \multirow{5}{*}{\rotatebox{90}{\textbf{\textit{Query}}}}&D. DETR-O~\cite{deformabledetr}   & R50   & 84.89 & 70.71 & 46.04 & 61.92 & 73.99 & 78.83 & 87.71 & 90.07 & 77.97 & 78.41 & 47.07 & 54.48 & 66.87 & 67.66 & 55.62 & 69.48 \\
          &EMO2-DETR~\cite{emo2detr}&R50 & 88.08 & 77.91 & 43.17 & 62.91 & 74.01 & 75.09 & 79.21 & 90.88 & 81.50 & 54.04 & 51.92 & 59.44 & 64.74 & 71.81 & 58.96 & 70.91  \\
          &ARS-DETR~\cite{arsdetr} & R50   & 86.97 & 75.56 & 48.32 & 69.20 & 77.92 & 77.94 & 87.69 & 90.50 & 77.31 & 82.86 & 60.28 & 64.58 & 74.88 & 71.76 & \dunderline{66.62} & 74.16 \\
          &RQFormer~\cite{RQFormer} & R50 & 87.45 & 78.57 & 47.36 & 69.01 & 79.58 & 81.27 & 88.53 & 90.89 & 82.80 & 86.21 & 58.68 & 64.20 & 75.21 & \dunderline{74.44} & 61.39 & 75.04 \\
          &O. Former~\cite{orientedformer}& R50 & 88.14 & 79.13 & 51.96 & 67.34 & 81.02 & 83.26 & 88.29 & 90.90 & 85.57 & 86.25 & 60.84 & 66.36 & 73.81 & 71.23 & 56.49 & 75.37  \\

         \midrule[0.8pt]

         \multirow{5}{*}{\rotatebox{90}{\textbf{\textit{GS}}}}&\textbf{GSDet (900 @ 1)} & R50 & 88.65 & 81.31 & 51.11 & 73.48 & 78.59 & 82.76 & 88.17 & 90.83 & 84.04 & 81.35 & 59.68 & 62.27 & 74.68 & 69.91 & 64.89 & 75.44 \\
         &\textbf{GSDet  (900 @ 3)} & R50 & 88.57 & 81.17 & 51.70 & 73.19 & 80.22 & 83.54 & \dunderline{88.64} & 90.82 & 84.10 & 81.59 & 59.48 & 64.69 & 75.08 & 70.00 & 63.33 & 75.74\\
         &\textbf{GSDet  (900 @ 1)} & Swin-T & 85.05 & 80.85 & 52.90 & 72.31 & 80.22 & \dunderline{83.65} & 87.58 & 89.65 & 83.95 & 85.65 & 62.15 & 62.94 & 76.28 & 72.19 & 61.64 & 75.80\\
         &\textbf{GSDet  (900 @ 3)} & Swin-T  & 85.89 & 81.41 & 52.98 & 72.52 & 80.59 & 83.12 & 87.96 & 90.04 & 83.92 & 85.86 & 62.51 & 63.13 & \dunderline{76.53} & 72.41 & 61.61 & \dunderline{76.03}\\
         &\textbf{GSDet  (900 @ 1)*} & Swin-T  & 88.04 & \textbf{85.55} & \textbf{57.22} & \textbf{79.44} & \textbf{81.28} & \textbf{84.77} & \textbf{88.73} & 90.82 & \dunderline{87.11} & \dunderline{87.32} & \textbf{68.64} & 68.54 & \textbf{79.57} & \textbf{80.81} & \textbf{72.79} & \textbf{80.04}\\
         
         \bottomrule[1.1pt]
        \end{tabular}
    }
    \caption{Comparison with state-of-the-art methods on \textbf{DOTA-v1.0}. * denotes multi-scale training and testing. (900@3) indicates that GSDet is evaluated with 900 Gaussians and 3 iterations in adaptive control. The \textbf{best} and \dunderline{second-best} results are highlighted in \textbf{bold} and \dunderline{underlined}.}
    \vspace{-0.5em}
    \label{tab:dota1.0}
    \end{table*}
}

\newcommand{\individual}{
    \begin{table}[t]
    \begin{minipage}[b]{0.49\linewidth}
      \centering
                \tablestyle{2.1pt}{1.12}
                \resizebox{1.01\linewidth}{!}
                {
                    \begin{tabular}{ccccc}
                    \toprule[1.02pt]
                    \multirow{2}{*}{\begin{tabular}[c]{@{}c@{}}\textbf{3D Gau.}\\ \textbf{Represent}\end{tabular}} & \multirow{2}{*}{\begin{tabular}[c]{@{}c@{}}\textbf{3D Gau.}\\ \textbf{Sample}\end{tabular}} & 
                    \multirow{2}{*}{\textbf{mAP}} & 
                    \multirow{2}{*}{\textbf{AP$_{75}$}} & 
                    \multirow{2}{*}{\textbf{AP$_{50}$}} \\
                               &            &       &       &       \\
                    \midrule[0.85pt]
                               &            & 38.96 & 39.03 & 67.01 \\
                    \checkmark &            & 47.03 & 50.61 & 74.01 \\
                    \checkmark & \checkmark & \textbf{47.77} & \textbf{52.15} & \textbf{75.44} \\
                    \bottomrule[1.02pt]
                    \end{tabular}
                }
      \caption{The effect of \textbf{3D Gaussian representation} and \textbf{sampling} on DOTA-v1.0.}\medskip
      \vspace{-1em}
      \label{table:individual_3D_gau}
    \end{minipage}
    \hfill
    \begin{minipage}[b]{0.49\linewidth}
      \centering
                \tablestyle{2.1pt}{1.12}
                \resizebox{1.01\linewidth}{!}
                {
                    \begin{tabular}{ccccc}
                    \toprule[1.02pt]
                    \multirow{2}{*}{\begin{tabular}[c]{@{}c@{}}\textbf{Self-}\\ \textbf{Attention}\end{tabular}} & \multirow{2}{*}{\begin{tabular}[c]{@{}c@{}}\textbf{Dynamic}\\ \textbf{Conv.}\end{tabular}} & 
                    \multirow{2}{*}{\textbf{mAP}} & 
                    \multirow{2}{*}{\textbf{AP$_{75}$}} & 
                    \multirow{2}{*}{\textbf{AP$_{50}$}} \\
                               &            &       &       &       \\
                    \midrule[0.85pt]
                               &            & 33.24 & 32.07 & 61.91 \\
                    \checkmark &            & 37.87 & 38.41 & 66.21 \\
                    \checkmark & \checkmark & \textbf{47.77} & \textbf{52.15} & \textbf{75.44} \\
                    \bottomrule[1.02pt]
                    \end{tabular}
                }
      \caption{The effect of \textbf{self-attention} and \textbf{dynamic convolution} on DOTA-v1.0.}\medskip
      \vspace{-1em}
      \label{table:individaual_attention}
    \end{minipage}
    \end{table} 
}

\newcommand{\individualplus}{
        \begin{table}[t]
    \begin{minipage}[b]{0.49\linewidth}
      \centering
                \tablestyle{2.1pt}{1.12}
                \resizebox{1.01\linewidth}{!}
                {
                    \begin{tabular}{lccc}
                    \toprule[1.02pt]
                    \textbf{Method} & \textbf{AP} & \textbf{AP$_{75}$} & \textbf{AP$_{50}$} \\
                    \midrule[0.85pt]
                    Cross-Attention & 38.22 & 38.42 & 69.29    \\
                    \textbf{Dynamic Conv.} & \textbf{47.77} & \textbf{52.15} & \textbf{75.44} \\
                    \bottomrule[1.02pt]
                    \end{tabular}
                }
      \caption{The comparison between \textbf{cross-attention} and \textbf{dynamic conv.} on DOTA-v1.0.}\medskip
            \vspace{-1em}
      \label{table:individual_cross_attn}
    \end{minipage}
    \hfill
    \begin{minipage}[b]{0.49\linewidth}
      \centering
                \tablestyle{2.1pt}{1.18}
                \resizebox{1.01\linewidth}{!}
                {
                    \begin{tabular}{lccc}
                    \toprule[1.01pt]
                    \textbf{Method} & \textbf{AP} & \textbf{AP$_{75}$} & \textbf{AP$_{50}$} \\
                    \midrule[0.8pt]
                    Feature pyramid & 45.13 & 47.17  & 73.21  \\
                    \textbf{3D feature space} & \textbf{47.77} & \textbf{52.15} & \textbf{75.44}   \\
                    \bottomrule[1.01pt]
                    \end{tabular}
                }
      \caption{The comparison between \textbf{feature pyramid} and \textbf{3D feature space} on DOTA-v1.0.}\medskip
            \vspace{-1em}
      \label{tab:individual_pyramid}
    \end{minipage}
    \end{table} 
}

\newcommand{\Initialization}{
        \begin{table}[!t]
    \centering
        \resizebox{0.99\linewidth}{!}
    {
        \begin{tabular}{cccccc}
        \toprule[1.01pt]
        \textbf{Metrics} & \textbf{Center} & \textbf{Full} & \textbf{Grid} & \textbf{Isotropic} & \textbf{Random} \\
        \midrule[0.8pt]
        mAP & 5.44 & 7.61 & 46.33 & 46.83 & 47.77 \\
        AP$_{75}$ & 4.06 & 6.13 & 50.85 & 51.12 & 52.15 \\ 
        AP$_{50}$ & 11.60 & 15.11 & 73.01 & 73.59 & 75.44 \\
        \bottomrule[1.01pt]
        \end{tabular}
    }
    \caption{Comparison with other \textbf{initialization} strategies on DOTA-v1.0. The random initialization works best.}
    \label{tab:initial}
    \end{table}
}

\newcommand{\adaptivecontrol}{
        \begin{table}[!t]
    \centering
        \resizebox{0.99\linewidth}{!}
    {
        \begin{tabular}{cccccccc}
        \toprule[1.01pt]
        \textbf{High-conf.} & \textbf{Low-conf.} &  & \textbf{AP$_{50}$} & \textbf{FPS} &  & \textbf{AP$_{50}$} & \textbf{FPS} \\ 
        \midrule[0.8pt]
                           &               & \multirow{4}{*}{\rotatebox{90}{\textbf{1 Iter.}}}  & 75.44        &  25  &  \multirow{4}{*}{\rotatebox{90}{\textbf{3 Iter.}}}   & 74.78   &  18       \\
        \checkmark         &               &   & 75.44        &  25  &     &   75.53  &  18        \\
                           & \checkmark    &   & 75.44        &  25  &     &   75.65  &  18        \\
        \checkmark         & \checkmark    &   & 75.44        &  25  &     &   75.74  &  18        \\ 
        \bottomrule[1.01pt]
        \end{tabular}
    }
    \caption{The effect of Gaussian update strategy in \textbf{adaptive control}.}
    \vspace{-1em}
    \label{tab:adaptive control}
    \end{table}
}

\usepackage{graphicx}

\newcommand{\teaser}{
    \begin{figure}[t]
      \centering
       \includegraphics[width=\linewidth]{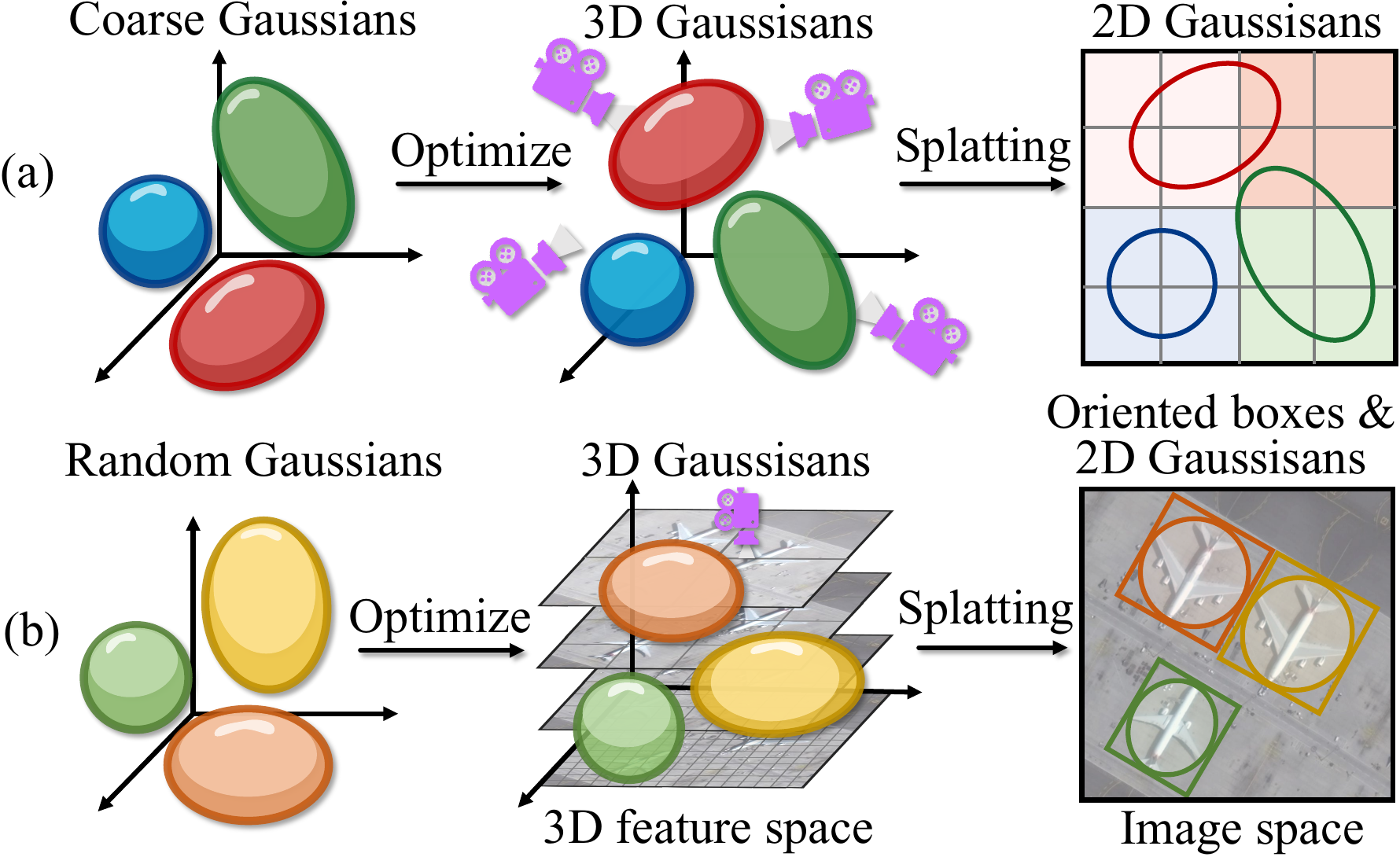}
       \caption{(a) \textbf{GS for rendering}. The 3D Gaussians are initialized through SFM or random points and optimized for mean, covariance, opacity, and color. All image pixels are computed through a tile-based rasterizer. (b) \textbf{GS for oriented object detection}. Randomly initialized 3D Gaussians are placed in a 3D feature space and trained to optimize mean, covariance, and confidence scores. The 3D Gaussians are then projected onto the image plane to form 2D Gaussians, which are subsequently transformed into oriented boxes.}
       \label{teaser}
       \vspace{-1em}
    \end{figure}
}

\newcommand{\teaserobject}{
    \begin{figure*}[!t]
    \centering
    \subfloat[\small Anchor]
        {
            \includegraphics[width=0.204\linewidth]{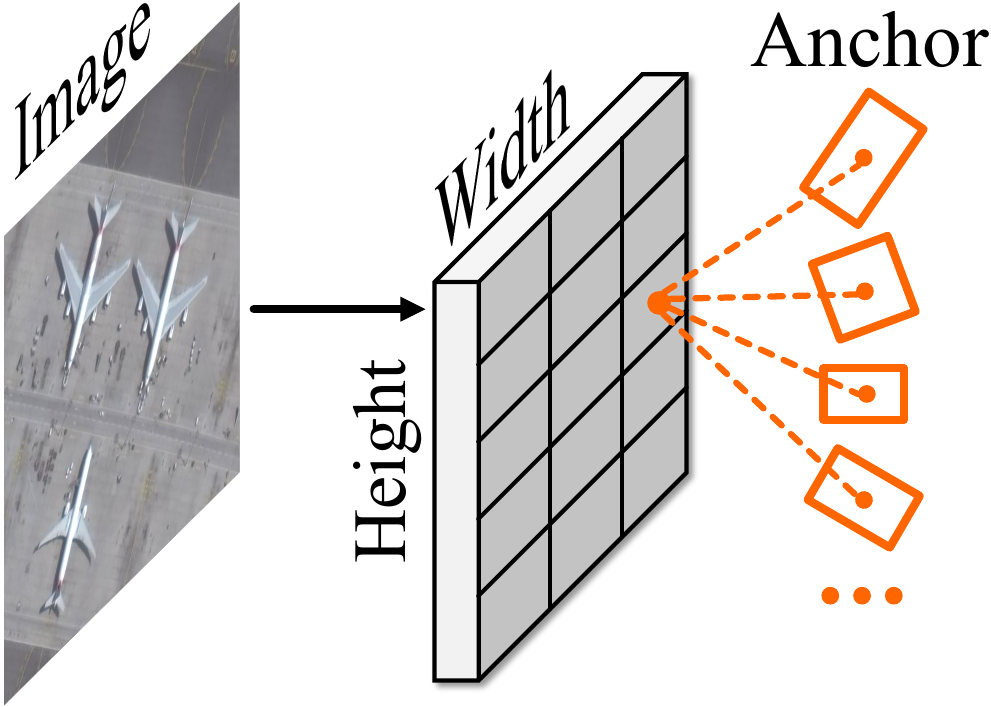}
            \label{pipeline_anchor}
        }
    \subfloat[\small Region proposal]
        {
            \includegraphics[width=0.204\linewidth]{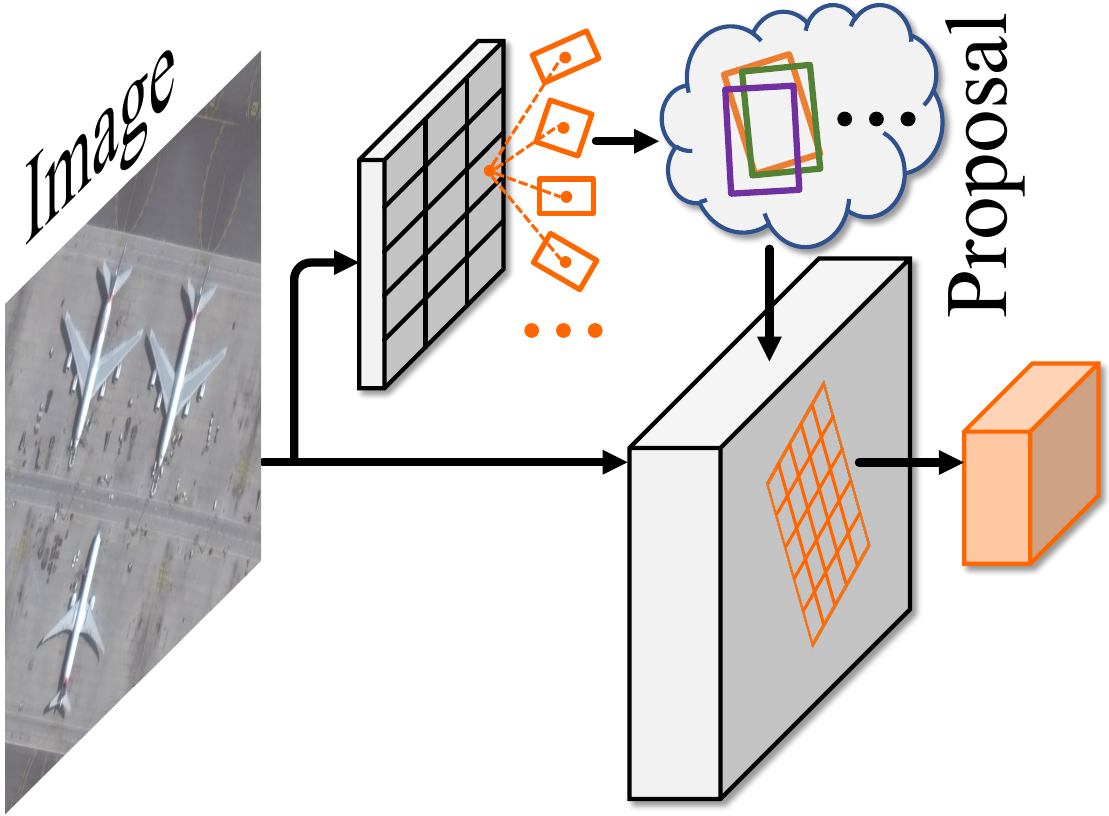}
            \label{pipeline_proposal}
        }
    \subfloat[\small Query]
        {
            \includegraphics[width=0.204\linewidth]{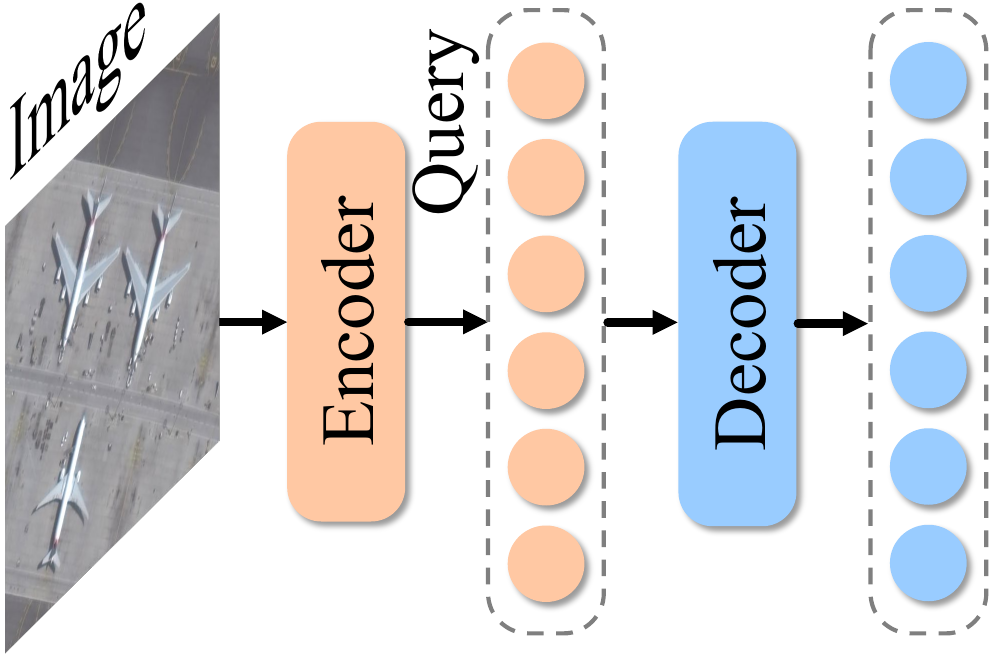}
            \label{pipeline_query}
        }
    \subfloat[\small Random Gaussian]
        {
            \includegraphics[width=0.344\linewidth]{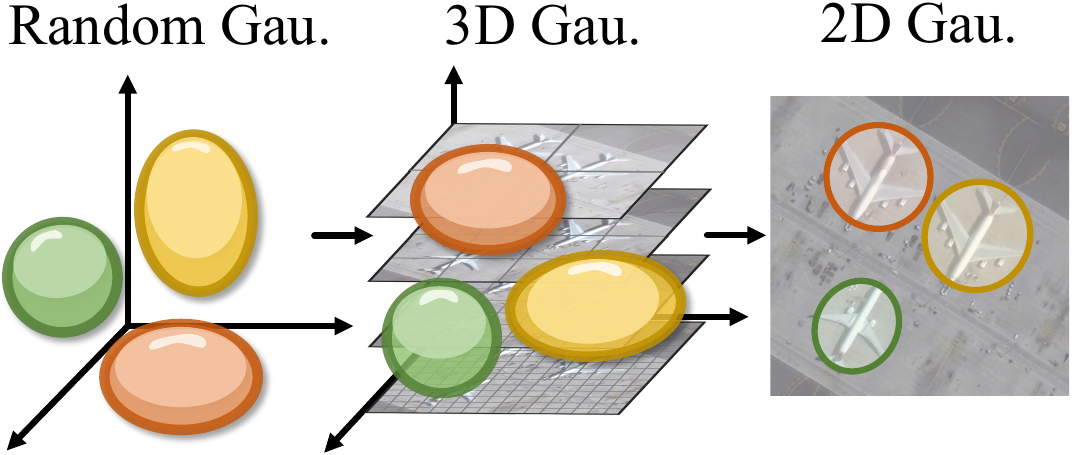}
            \label{pipeline_gs}
        }
    \caption{Comparisons of \textbf{different schemes for object candidates}. (a) In one-stage CNN-based detectors, dense anchors are enumerated across all image grids. (b) In two-stage CNN-based detectors, region proposals are selected through region proposal networks. (c) In transformer-based detectors, learnable queries are produced by the encoder. (d) In our GSDet, 3D Gaussians are randomly initialized.}
    \label{teaserobject}
    \vspace{-1em}
    \end{figure*}
}

\newcommand{\overview}{
    \begin{figure*}[t]
      \centering
       \includegraphics[width=0.88\textwidth]{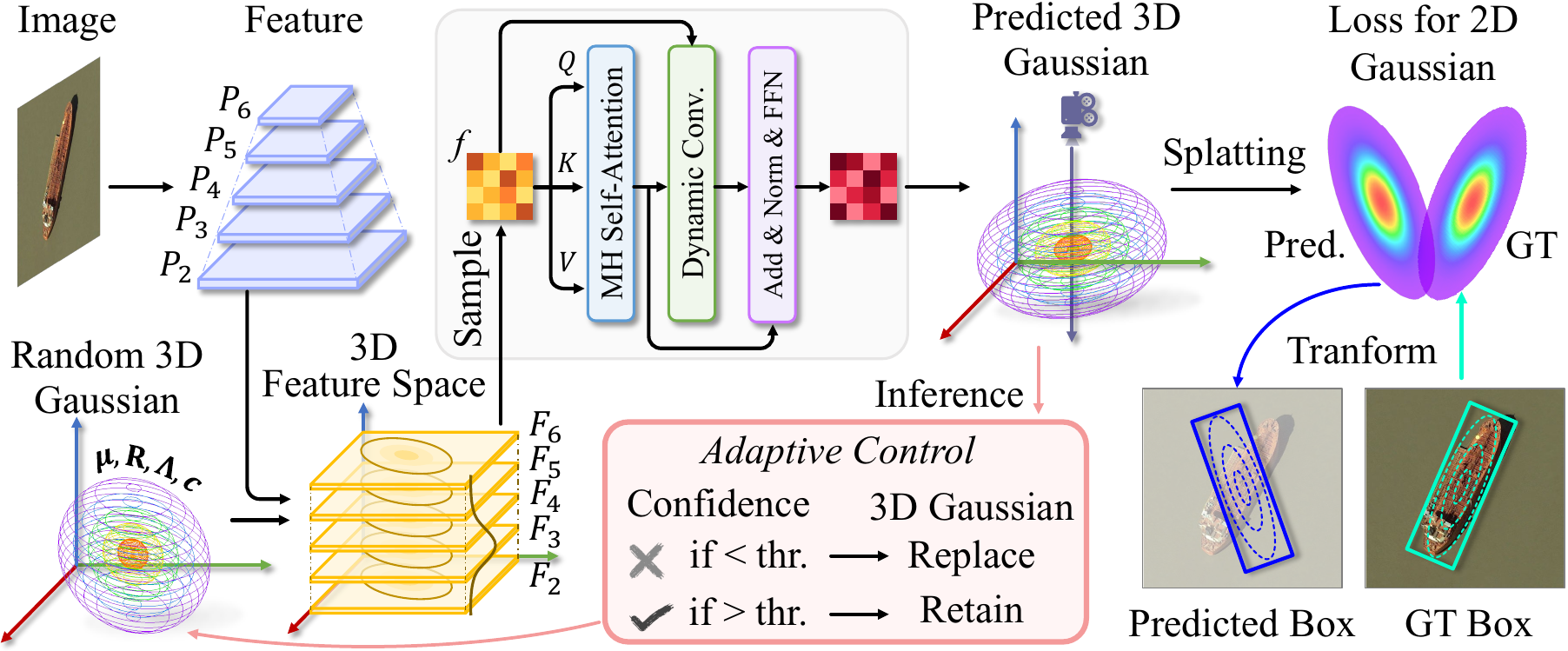}
       \caption{\textbf{The overview of GSDet}. 3D Gaussians, defined by their mean, covariance, and confidence scores, are employed to represent objects and are initialized randomly. A 3D feature space is constructed from image features, where 3D Gaussian sampling is designed to capture spatial information. The decoder outputs predicted 3D Gaussians, which are projected into 2D Gaussians, supervised via loss functions, and ultimately transformed into oriented boxes. During inference, adaptive control and a dynamic number of Gaussians are introduced.}
       \label{overview}
    \end{figure*}
}

\newcommand{\dynamicnumberofgau}{
    \begin{figure}[t]
       \centering
       \includegraphics[width=0.93\linewidth]{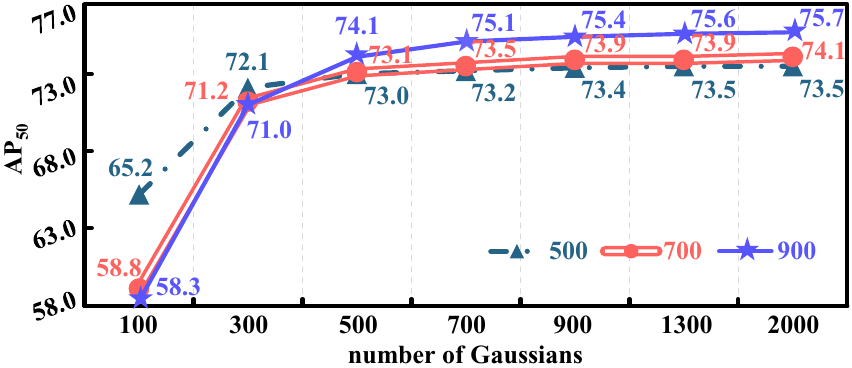}
       \vspace{-0.5em}
       \caption{\textbf{Dynamic number of Gaussians} during inference.}
       \vspace{-0.5em}
       \label{dynamic_number_of_gau}
    \end{figure}
}

\newcommand{\iterativeevaluation}{
    \begin{figure}[t]
    \begin{minipage}[b]{0.48\linewidth}
      \centering
    \centerline{\includegraphics[width=1\linewidth]{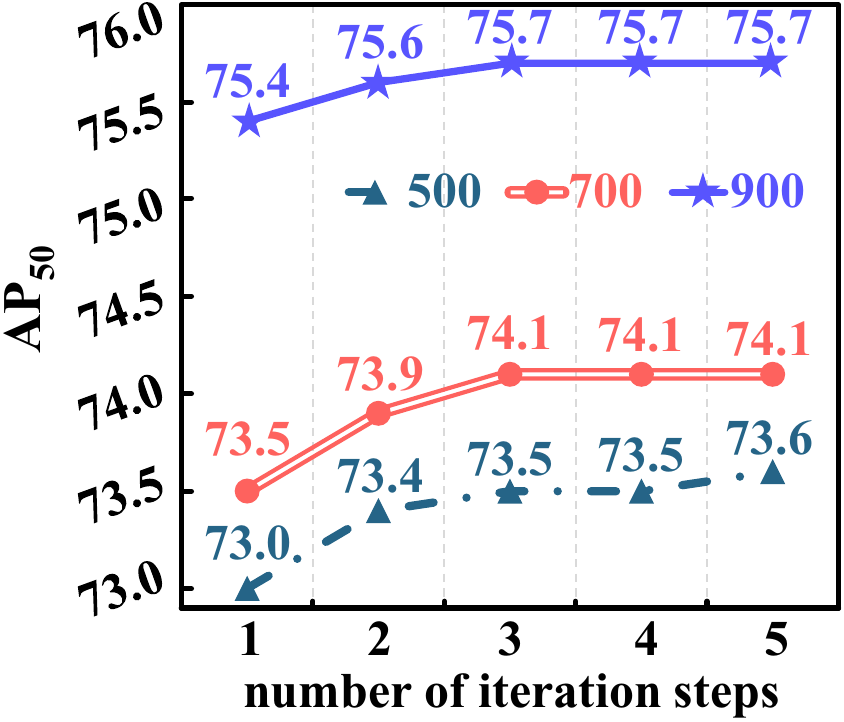}}
    \vspace{-0.5em}
      \caption{The dynamic number of iterations in \textbf{adaptive control} during inference.}\medskip
      \vspace{-1em}
      \label{fig:iter_step}
    \end{minipage}
    \hspace{0.08em}
    \begin{minipage}[b]{0.485\linewidth}
     \centering
    \centerline{\includegraphics[width=1\linewidth]{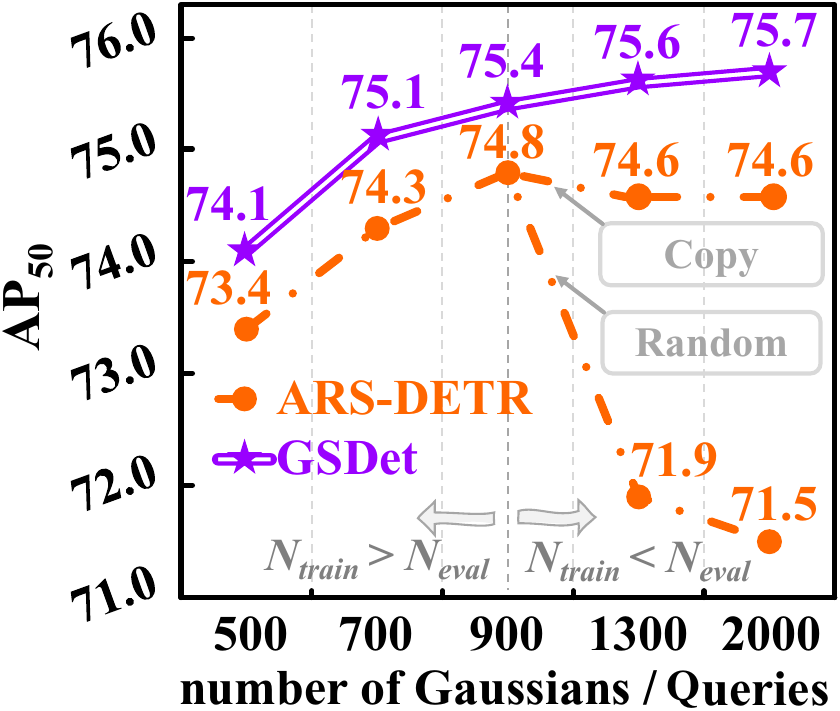}}
    \vspace{-0.5em}
      \caption{\textbf{GSDet vs. ARS-DETR} with dynamic number of Gaussians during inference.}\medskip
      \vspace{-1em}
      \label{fig:GSDet_vs_ARSDETR}
    \end{minipage}
    \end{figure} 
}

\newcommand{\visgau}{
    \begin{figure}[!t]
      \centering
       \includegraphics[width=0.99\linewidth]{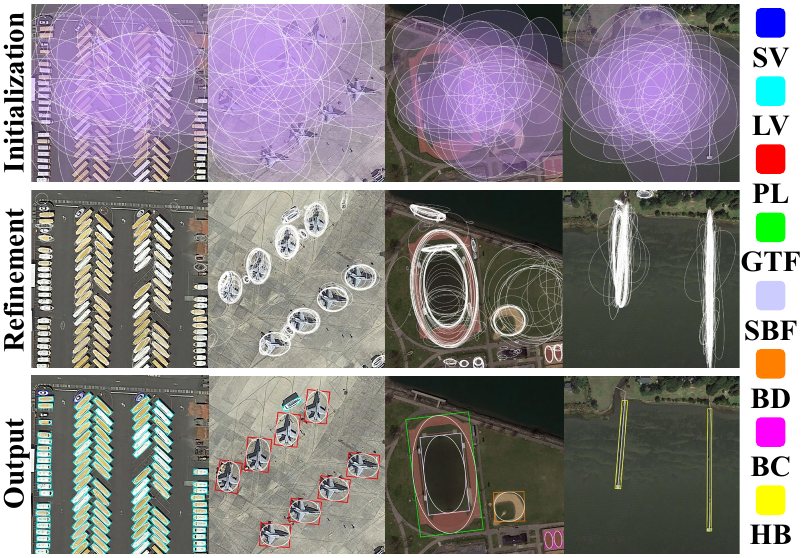}
       \caption{\textbf{Top:} random Gaussians. Only a small number of random 2D Gaussians are shown for better visualization. \textbf{Middle:} refined Gaussians from the middle decoder layer focus on foreground objects. \textbf{Bottom:} final Gaussians and predicted results.}
       \label{fig:vis_gau}
    \end{figure}
}

\newcommand{\visdiffinput}{
    \begin{figure}[!t]
      \centering
       \includegraphics[width=0.95\linewidth]{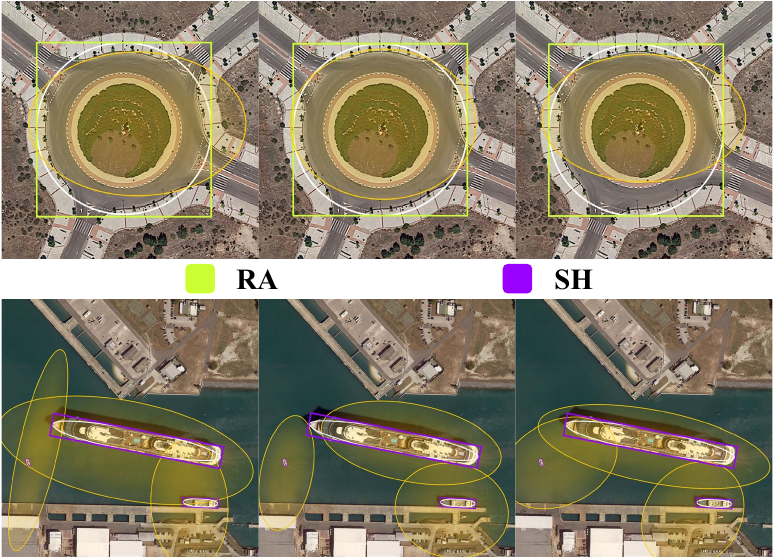}
       \caption{Different randomly initialized Gaussians with a yellow color and corresponding predicted results.}
       \label{fig:vis_diff_input}
    \end{figure}
}

\begin{document}

\maketitle

\begin{abstract}
        Oriented object detection has advanced with the development of convolutional neural networks (CNNs) and transformers. However, modern detectors still rely on predefined object candidates, such as anchors in CNN-based methods or queries in transformer-based methods, which struggle to capture spatial information effectively. To address the limitations, we propose GSDet, a novel framework that formulates oriented object detection as Gaussian splatting. Specifically, our approach performs detection within a 3D feature space constructed from image features, where 3D Gaussians are employed to represent oriented objects. These 3D Gaussians are projected onto the image plane to form 2D Gaussians, which are then transformed into oriented boxes. Furthermore, we optimize the mean, anisotropic covariance, and confidence scores of these randomly initialized 3D Gaussians, using a decoder that incorporates 3D Gaussian sampling. Moreover, our method exhibits flexibility, enabling adaptive control and a dynamic number of Gaussians during inference. Experiments on 3 datasets indicate that GSDet achieves AP$_{50}$ gains of 0.7\% on DIOR-R, 0.3\% on DOTA-v1.0, and 0.55\% on DOTA-v1.5 when evaluated with adaptive control and outperforms mainstream detectors. Code link \url{https://github.com/wokaikaixinxin/GSDet}.
\end{abstract}

\section{Introduction}

Oriented object detection aims to identify and locate objects with arbitrary orientations in images~\cite{GauCho,orientedrcnn}. Unlike objects in horizontal object detection, oriented objects often appear in various scales, aspect ratios, and orientations, which significantly increases the difficulty of accurate detection~\cite{dior,dotav1.0}. These challenges necessitate the development of techniques that can effectively localize and classify oriented objects regardless of their complexity.

\teaser

Over the years, oriented object detection has evolved significantly, driven by advancements in both CNN-based~\cite{orientedrcnn,s2anet} and transformer-based~\cite{arsdetr} methods. CNN-based object detectors can be broadly categorized into one-stage and two-stage detectors. One-stage detectors~\cite{s2anet} predict object categories and bounding boxes directly from the feature map in a single pass, utilizing dense object candidates like anchors to cover potential object locations. In contrast, two-stage detectors~\cite{orientedrcnn} first generate region proposals using a region proposal network and then refine these proposals to produce the final results. Recently, transformer-based methods~\cite{arsdetr} have gained traction in this field, which treat object detection as a set prediction problem and use a set of learnable queries from the encoder to directly predict the final bounding boxes and class labels.

\teaserobject

Although CNN-based and transformer-based methods have demonstrated promising performance, they still rely on manually designed object candidates, such as anchors, region proposals, and queries. Moreover, these candidates, which represent potential objects, struggle to effectively capture spatial information of objects.

To address these issues, we propose a novel framework that directly detects oriented objects from random 3D Gaussians. The random initialization does not involve learnable neuron parameters. Regarding spatial information, a 3D feature space is constructed from image features. Within this space, the 3D Gaussians are optimized to represent oriented objects and can be projected onto the image plane to form 2D Gaussians, which are then transformed into oriented boxes. This \textit{Gaussian-to-box} method eliminates the need for prior anchors, proposals, or queries, which simplifies the generation of object candidates and establishes a streamlined pipeline.

Our method is inspired by Gaussian splatting (GS), as illustrated in Figure~\ref{teaser}. The philosophy of \textit{Gaussian-to-box} paradigm is analogous to the \textit{Gaussian-to-image} process in Gaussian splatting~\cite{gsrender}, where a scene is represented by 3D Gaussians and high-quality rendering is achieved by projecting these 3D Gaussians onto the 2D image space. Gaussian splatting has demonstrated significant advancements in 3D scene representation tasks. However, to the best of our knowledge, no prior work has successfully applied Gaussian splatting to oriented object detection.

In this paper, we propose GSDet, a novel approach that formulates oriented object detection as Gaussian splatting. 
A 3D feature space is constructed by treating multi-scale feature maps from the encoder as supporting planes within a 3D coordinate system. In this space, 3D Gaussians serve as a flexible representation of objects and are initialized randomly, without relying on structure-of-motion (SFM) points. The Gaussian attributes, including the mean, anisotropic covariance, and confidence scores, are optimized using a decoder that incorporates 3D Gaussian sampling to extract spatial features. The predicted 3D Gaussians are projected onto the image plane to form 2D Gaussians, which are further transformed into predicted boxes via decomposition.

Our GSDet exhibits remarkable flexibility during inference. Firstly, it allows for a dynamic number of initial 3D Gaussians as input. We can train the model with a certain number of Gaussians and then use a different number during inference. Secondly, GSDet enables iterative reuse of the decoder to refine the predicted Gaussians through adaptive control, allowing Gaussians to focus on foreground objects.

Experiments on 3 datasets validate the effectiveness of GSDet. Its performance is further improved through the use of adaptive control and an increased number of Gaussians during inference. Contributions are summarized as follows:
\begin{itemize}
\item We propose a detector GSDet, which is the first work to formulate oriented object detection as Gaussian splatting to the best knowledge of our knowledge.
\item We perform detection within a 3D feature space constructed from image features, where the designed 3D Gaussian sampling captures spatial information.
\item Anisotropic 3D Gaussians are employed to represent oriented objects through their attributes. They are projected onto the image plane to form 2D Gaussians, which are then transformed into oriented boxes.
\item Our Gaussian-to-box detection paradigm exhibits favorable properties, including adaptive control and dynamic Gaussians during inference, owing to the random initialization of 3D Gaussians.
\end{itemize} 

\overview

\section{Related Work}

\textbf{Oriented object detection.} The mainstream methods in oriented object detection can be broadly categorized into CNN-based and transformer-based approaches. CNN-based methods, which have achieved significant success, are further divided into single-stage and two-stage methods. Two-stage methods~\cite{roi_transformer,orientedrcnn} utilize a region proposal network (RPN) to generate dense proposals, which are then refined by a region of interest (RoI) head. In contrast, single-stage methods~\cite{sasm,dhrec} bypass the need for region proposals and detect objects directly from dense anchors.

To eliminate the need for dense anchor or proposal candidates, transformer-based methods establish a paradigm based on a set of queries~\cite{arsdetr,emo2detr}. These queries represent both the semantic and location information of objects, and are selected through the encoder~\cite{deformabledetr}. Mainstream transformer architectures consist of an encoder and a decoder, with attention mechanisms serving as their core. Self-attention in the decoder enables queries to interact with one another, while cross-attention integrates encoder features with queries.

Unlike these methods, our approach is the first to predict oriented boxes directly from random Gaussians using Gaussian splatting, eliminating the need for pre-designed anchors, proposals, or learned queries, as illustrated in Figure~\ref{teaserobject}.

\textbf{Gaussian splatting} has emerged as a significant technique in computer graphics, offering explicit scene representation and efficient rendering capabilities~\cite{gsrender}. It represents objects and surfaces as collections of Gaussians, allowing for high-quality image synthesis. This method has been explored in various applications, including scene reconstruction~\cite{deformable3dgau,4dgausplat}, SLAM~\cite{gsslam,matsuki2024gaussian}, and autonomous driving~\cite{drivinggaussian,hugs}, demonstrating its versatility and potential for real-time rendering. Despite the success of Gaussian splatting in these domains, there are no previous solutions that successfully adapt Gaussian splatting for oriented object detection. We believe that the slow progress of Gaussian splatting in oriented object detection is primarily due to the different optimization objectives, \emph{i.e.}, Gaussians and oriented boxes. To the best of our knowledge, this is the first work that adopts Gaussian splatting for oriented object detection.


\section{Method}
Oriented object detection identifies oriented boxes and class labels for objects in images. The core of our GSDet is to address oriented object detection through Gaussian splatting. In this section, we provide a detailed description of our detector. The overview of our GSDet is illustrated in Figure~\ref{overview}.

\subsection{Gaussian for Oriented Object}\label{gau_rep}
\textbf{Oriented box.} The objects rotate arbitrarily in oriented object detection. Typically, an oriented box is represented as $(cx, cy, w, h, \theta)$, where $cx$ and $cy$ denote the center coordinates, $w$ and $h$ represent the width and height, and $\theta$ specifies the angle of orientation. To make better use of spatial features, we introduce another representation $(cx, cy, cz, s, \theta)$, where $cz$ denotes the scale of objects and $s$ denotes the aspect ratio. The $cz$ and $s$ are calculated as:
\begin{equation}
    \label{new_rep_box}
  cz = \log _2(\sqrt{wh}),s=\log_2(\sqrt{\frac{h}{w}}),
\end{equation}
where $wh$ denote the area, and $\frac{h}{w}$ denote the aspect ratio of objects. The $\log_2(\cdot)$ is used to limit the range of value. Furthermore, the $(cx, cy, cz)$ directly represents the center of a 3D Gaussian in a 3D space, which will be introduced below.

\textbf{3D Gaussian representation.} In our approach, anisotropic 3D Gaussians are initialized randomly. The basic 3D Gaussian $\mathcal{N}_{3d}(\boldsymbol{\mu}_{3d},\space \boldsymbol{\Sigma}_{3d})$ is described by its mean $\boldsymbol{\mu}_{3d}$, covariance matrix $\boldsymbol{\Sigma}_{3d}$, and confidence score $\boldsymbol{c}$.

Geometrically, the mean $\boldsymbol{\mu}_{3d}$ represents the center coordinates of ellipsoids, and can be defined as:
\begin{equation}
  \boldsymbol{\mu}_{3d} = (cx,cy,cz)^\top.
\end{equation}

The covariance matrices $\boldsymbol{\Sigma}_{3d}$ have geometric meaning only when they are positive semi-definite. Typically, it is difficult to constrain the learnable model parameters using gradient descent to generate such valid matrices. To avoid invalid matrices during training, we optimize the factorized form of the covariance matrix. The covariance matrix can be factorize into a rotation matrix $\mathbf{R}_{3d}$ and scaling matrix $\mathbf{\Lambda}_{3d}$:
\begin{equation}
    \boldsymbol{\Sigma}_{3d}=(\mathbf{R}_{3d}\mathbf{\Lambda}_{3d})(\mathbf{R}_{3d}\mathbf{\Lambda}_{3d})^\top,
\end{equation}
where the rotation matrix $\mathbf{R}_{3d}$ and scaling matrix $\mathbf{\Lambda}_{3d}$ are expressed as:
\begin{equation}\label{r3d_lambda3d}
    \hspace{-0.5em}
  \mathbf{R}_{3d}=\begin{pmatrix}
  \cos\theta&  -\sin\theta & 0\\
  \sin\theta& \cos\theta & 0\\
  0& 0 & 1
\end{pmatrix},
\mathbf{\Lambda}_{3d}=\begin{pmatrix}
  \lambda _1&  & \\
  & \lambda _2 & \\
  &  &\lambda _3
\end{pmatrix}.
\end{equation}
The rotation matrix $\mathbf{R}_{3d}$ describes the rotation of the entire Gaussian by angle $\theta$ around a specified axis geometrically. In our setting, the 3D Gaussians rotate only around the $z$-axis, as the image is captured from a single perspective. The scaling matrix $\mathbf{\Lambda}_{3d}$ changes the size of a Gaussian by scaling it along the $x$-, $y$-, and $z$-axis. The $\lambda_1$, $\lambda_2$, and $\lambda_3$ are scaling factors in different eigenvector directions and determine how much the Gaussian is stretched or shrunk in each direction. Specifically, the $\lambda_1$ and $\lambda_2$ are calculated as:
\begin{equation}
    \lambda _1=\frac{1}{2\sigma }\cdot 2^{cz-s}, \; \lambda _2=\frac{1}{2\sigma }\cdot 2^{cz+s},
    \label{lambda1and2}
\end{equation}
where $\sigma=3$ by default according to the 3-sigma rule. The $\lambda _3$ is not explicitly defined, but features closer to $cz$ receive more attention, which will be elaborated in Sec.~\ref{architecture}.

The confidence score $\boldsymbol{c}$ is used for classification.

\textbf{Splatting: 3D Gaussian $\rightarrow$ 2D Gaussian.} In oriented object detection, only the original image view is considered, not new viewpoints. Consequently, the 3D Gaussians are projected onto the image plane along the $z$-axis to produce 2D Gaussians $\mathcal{N}_{2d}(\boldsymbol{\mu}_{2d},\space \boldsymbol{\Sigma}_{2d})$. The mean $\boldsymbol{\mu}_{2d}$ is calculated as:
\vspace{-0.2em}
\begin{equation}
  \boldsymbol{\mu}_{2d}=(cx, cy)^\top.
  \label{mu2d}
  \vspace{-0.2em}
\end{equation}

The covariance $\boldsymbol{\Sigma}_{2d}$ can also be factorized by rotation matrix $\mathbf{R}_{2d}$ and scaling matrix 
$\mathbf{\Lambda}_{2d}$, \emph{i.e.}, $\boldsymbol{\Sigma}_{2d}=(\mathbf{R}_{2d}\mathbf{\Lambda}_{2d})$ $(\mathbf{R}_{2d}\mathbf{\Lambda}_{2d})^\top$. The $\mathbf{R}_{2d}$ and $\mathbf{\Lambda}_{2d}$ can be calculated as:
\vspace{-0.2em}
\begin{equation}
  \mathbf{R}_{2d}=
\begin{pmatrix}
 \cos \theta  &-\sin \theta  \\
 \sin \theta  &\cos \theta 
\end{pmatrix},
\mathbf{\Lambda}_{2d}=
\begin{pmatrix}
  \lambda _1& \\
  &\lambda _2
\end{pmatrix}.
\label{r2d}
\vspace{-0.2em}
\end{equation}

The confidence score $\boldsymbol{c}$ is consistent between each 2D Gaussian and its corresponding 3D Gaussian.

The time consumption in the splatting process is negligible, as 3D Gaussians can directly achieve orthogonal projection using these concise formulas.

\textbf{Transformation: 2D Gaussian $\rightarrow$ Oriented box.}
Each Gaussian characterizes a potential object in our method, rather than multiple Gaussians jointly determining a potential object. Thus, given the mean $\boldsymbol{\mu}_{2d}$, rotation matrix $\mathbf{R}_{2d}$, and scaling matrix $\mathbf{\Lambda}_{2d}$, the oriented boxes as final results are transformed from 2D Gaussians by the reverse process of Eq.~\ref{new_rep_box}, Eq.~\ref{lambda1and2}, Eq.~\ref{mu2d} and Eq.~\ref{r2d} through decomposition.

\subsection{Architecture}
\label{architecture}
Our framework consists of an image encoder and a detection decoder, which is a simple paradigm. The image encoder provides 3D feature space and the detection decoder is trained to refine the random 3D Gaussians.

\textbf{Image encoder.} The image encoder is composed of a backbone~\cite{resnet,swin} and a feature pyramid network (FPN)~\cite{fpn}. The backbone takes in the raw image as its input and extracts image features for the subsequent FPN. The FPN is utilized to generate multi-scale feature maps with 256 channels. Specifically, we construct the pyramid with levels ranging from $P_2$ to $P_6$, where $P_l$ has a resolution that is $2^l$ times lower than that of the image. The $l\in\{2,3,4,5,6\}$ represents the pyramid level.

\textbf{3D feature space.} Since oriented object detection lacks real depth information, we construct a 3D feature space based on the feature pyramid. 
The $z$-axis coordinates $Z_l$ of each feature map are computed as:
\vspace{-0.2em}
\begin{align}
    Z_l = \log_{2}(2^l/4).
\vspace{-0.2em}
\end{align}
We rescale the feature maps $P_3\sim P_6$ to the same size $[W_F,H_F]$ as $P_2$ by interpolation.
These feature maps, acting as supporting planes, are aligned along the $x$- and $y$-axes in the 3D feature space, denoted as $\{F_2, F_3, F_4, F_5, F_6\}$.

\textbf{3D Gaussian sampling.}
We implement 3D Gaussian sampling in a decomposition manner. We first sample in the feature maps and then integrate the Gaussian weights of the $z$-axis. The main sampling area $G^{xy}(i,j)$ based on the 3-sigma rule for each feature map is calculated as follows:
\vspace{-0.2em}
\begin{equation}
    \texttt{sum}(((\sigma\mathbf{\Lambda}_{2d} )^{-1}\mathbf{R}_{2d}((i,j)^\top-\boldsymbol{\mu}_{2d}))^{\circ 2 })\le 1,
\end{equation}
where $(i,j)\in[0,W_F]\times[0,H_F]$. $(\cdot )^{\circ 2}$ denotes squaring each element. $\texttt{sum}(\cdot)$ denotes the summation of all elements. The Gaussian weight on the $z$-axis is calculated as follows:
\begin{equation}\label{gau_weight}
  G^z_l=\frac{\exp(-(Z_l-cz)^2)}{\Sigma_l\exp(-(Z_l-cz)^2)}.
\end{equation}
The sampling area $G^{xy}$ and weight $G^z_l$ assigns high importance to spatial locations near the center $(cx, cy, cz)$. The 3D Gaussian sampling is calculated as follows:
\begin{equation}
  f_l=G^z_l\cdot \texttt{GauAlign}(G^{xy}, F_l),
\end{equation}
where $f_l$ denotes the sampled features from $F_l$. Inspired by $\texttt{RRoIAlign}()$~\cite{roi_transformer}, we introduce $\texttt{GauAlign}()$, which first crops main regional features based on $G^{xy}$ and then resizes them into a uniform size of $7\times 7$ with 256 channels using bilinear interpolation. This operator naturally aligns the oriented Gaussians with axis-aligned features. Last, the sampled features $f$ are obtained by summing $f_l$ from all levels, calculated as $f={\textstyle \sum_{l=2}^{6}}f_l $.

\textbf{Detection decoder.} The decoder stacks multiple layers, each integrating 3D Gaussian sampling, self-attention~\cite{detr}, and dynamic convolution~\cite{sparsercnn}. The $f$ sampled through 3D Gaussian sampling is passed into self-attention for the relations between Gaussians. Dynamic convolution further enhances feature fusion. Finally, a feedforward network outputs 3D Gaussians as predictions.

\GSDettrain

\subsection{Training}\label{training}
The training procedure for GSDet is shown in Algorithm~\ref{alg:gsdet_train}. Given input images and ground truth (GT), the model is trained to predict both locations and classes of objects.

\textbf{Loss for 2D Gaussians.} We assign multiple predictions to GT and apply the set prediction loss~\cite{detr}. We measure the distance and overlap between the predicted and GT 2D Gaussians. The GT 2D Gaussians are transformed from GT annotations. Specifically, the Gaussian Wasserstein distance (gwd) is employed to quantify the distance~\cite{gwd}, and the SkewIoU is used to calculate the overlap~\cite{kfiou}. Additionally, Focal loss~\cite{focal_loss} is used for classification. The loss is formulated as:
\begin{equation}
  \mathcal{L}=\delta_{gwd}\mathcal{L}_{gwd}+\delta _{skewiou}\mathcal{L}_{skewiou}+\delta_{cls}\mathcal{L}_{cls},
\end{equation}
where $\delta_{gwd}$, $\delta _{skewiou}$ and $\delta_{cls}$ are weights of each component. We set $\delta_{gwd}=2$, $\delta _{skewiou}=5$, and $\delta_{cls}=2$ by default.

\subsection{Inference}\label{inference}
\textbf{Adaptive control of Gaussian.}
We can evaluate GSDet with a dynamic number of iterations by reusing the decoder during inference. After each iteration, some predicted 3D Gaussians focus on foreground objects, while others are located in the background. Predicted Gaussians with lower confidence scores are replaced with random Gaussians for the next iteration. Meanwhile, high-confidence Gaussians are retained for final results and then replaced with relatively random Gaussians centered within their areas for the next iteration.

\textbf{Dynamic number of Gaussian.}
Thanks to the random initialization, we can evaluate GSDet with a dynamic number of random Gaussians during inference which do not need to be equal to the training stage. In contrast, previous methods rely on the same number of anchors or queries during training and inference, which is inflexible.

\section{Experiments}
\iterativeevaluation
\dynamicnumberofgau
\DOTAOneResults
\DIORRResults

\subsection{Experimental Setups}
\textbf{Datasets.} We conduct extensive experiments on three datasets DOTA-v1.0~\cite{dotav1.0}, DOTA-v1.5~\cite{dotav1.0} and DIOR-R~\cite{dior}.
\begin{itemize}
    \setlength{\itemsep}{-0.2em}    
    \item \textbf{DOTA-v1.0}~\cite{dotav1.0} comprises 1,869 images in the \texttt{trainval} set and 937 images in the \texttt{test} set, annotated with 188,282 instances across 15 categories.
    \item \textbf{DOTA-v1.5}~\cite{dotav1.0} dataset extends the DOTA-v1.0 dataset by adding a new category named ``Container Crane" while keeping the same images. The number of instances is increased to 403,318 in total.
    \item \textbf{DIOR-R}~\cite{dior} dataset consists of 11,725 training images in the \texttt{trainval} set, 11,738 test images in the \texttt{test} set and 192,512 instances belonging to 20 categories.
\end{itemize}

\textbf{Implementation details.} Our code is built on MMrotate with pytorch. The optimizer AdamW~\cite{adamw} is used with the learning rate as $2.5\times10^{-5}$ and the weight decay as $10^{-4}$. All models are trained with the batchsize 4 on two Nvidia 2080ti (2 images per GPU). The training schedule is 24 epochs, with the learning rate divided by 16 and 22 epochs. Data augmentation strategies contain only random flips. For DOTA-v1.0 and DOTA-v1.5 datasets, images are cropped into patches of $1024\times1024$ with overlaps of 200. The predictions of DOTA-v1.0/v1.5 are submitted to the official website for evaluation.

\textbf{Evaluation metrics.} The average precision at various intersections over union (IoU) thresholds is utilized to measure the precision and recall rates of the detected objects, including AP$_{50}$, AP$_{75}$, and mAP. Additionally, frames per second (FPS) is employed to evaluate the speed of methods.

\vspace{-0.2em}
\subsection{Properties of GSDet}
\textbf{Adaptive control.} We investigate the effect of adaptive control by increasing the number of iterations from 1 to 5 during inference, as shown in Figure~\ref{fig:iter_step}. The model is trained with 500, 700, and 900 Gaussians, respectively. As the number of iterations increases from 1 to 5, the model achieves a 0.3\% AP$_{50}$ improvement with 900 training Gaussians and a 0.6\% AP$_{50}$ improvement with both 500 and 700 training Gaussians.

\textbf{Dynamic number of Gaussians.} Figure~\ref{fig:GSDet_vs_ARSDETR} compares our GSDet with ARS-DETR~\cite{arsdetr}. Both models are trained with 900 Gaussians or queries and evaluated using {500, 700, 900, 1300, 2000} Gaussians or queries. To enable ARS-DETR to evaluate with 1300 or 2000 queries, we employ two strategies: cloning existing queries with non-maximum suppression and concatenating random queries. With the former strategy, ARS-DETR experiences a slight decrease in accuracy, whereas the latter strategy results in a significant accuracy drop. In contrast, as the number of Gaussians increases from 500 to 2000 during inference, GSDet achieves a 1.6 AP$_{50}$ improvement.

Additionally, we train GSDet with 500, 700, and 900 Gaussians and evaluate its performance across varying numbers of Gaussians during inference. Figure~\ref{dynamic_number_of_gau} illustrates that as the number of Gaussians increases from 100 to 2000, the AP$_{50}$ consistently improves, demonstrating the flexibility.

\Initialization

\vspace{-0.2em}
\subsection{Main Results}
We compare GSDet with modern CNN-based one-stage, two-stage, and transformer-based detectors. We train and evaluate GSDet with 900 Gaussians in both Table~\ref{tab:dota1.0} and~\ref{tab:diorr}. Table~\ref{tab:dota1.0} presents the detailed results for each category on DOTA-v1.0. GSDet achieves an AP$_{50}$ of 75.44\% with one iteration and 75.74\% with three iterations using the ResNet50~\cite{resnet}. Notably, its performance improves further when using the Swin-T~\cite{swin}. Furthermore, with a multi-scale training strategy, GSDet achieves an AP$_{50}$ of 80.04\%.

Table~\ref{tab:diorr} provides the AP$_{50}$, AP$_{75}$, mAP, and FPS results on DIOR-R, DOTA-v1.0, and DOTA-v1.5. GSDet achieves a AP$_{50}$ of 70.05\% on DIOR-R, 75.74\% on DOTA-v1.0, and 67.84\% on DOTA-v1.5 with three iterations. These results confirm that Gaussian splatting for oriented object detection outperforms both CNN-based and transformer-based methods. Additionally, the inference speed of GSDet is faster than that of two-stage and transformer-based methods owing to random initialization and simple architecture.

\individual
\vspace{-0.2em}
\subsection{Ablation Study}

\textbf{Gaussian initialization.} We study the effect of different Gaussian initialization strategies in Table~\ref{tab:initial}, including: (1) ``Center'' means all Gaussians are located in the central region, with a height and width set to 10\% of the image size; (2) ``Full'' indicates that all Gaussians are initialized to cover the entire image; (3) ``Grid'' means that each Gaussian is distributed within each grid, with a total of $30\times30$ grids arranged across the image; (4) ``Isotropic'' signifies that all Gaussians are isotropic and randomly initialized. The performance of the ``Center'' and ``Full'' strategies is notably poor, with AP$_{50}$ of only 11.60\% and 15.11\%, respectively. We argue that Gaussians under these two strategies can not effectively cover potential object regions. In contrast, the ``Random'' initialization achieves the best performance, with an AP$_{50}$ of 75.44\%, followed by the ``Isotropic'' and ``Grid'' strategies. The results indicate that anisotropic random Gaussians are more likely to align with potential object areas.

\textbf{3D Gaussian representation and sampling.} We evaluate the impact of 3D Gaussian representation and 3D Gaussian sampling in Table~\ref{table:individual_3D_gau}. This ablation begins with an additional angle branch~\cite{csl} and a set of randomly initialized rectangles, achieving an AP$_{50}$ of 67.01\%. Under these conditions, learning object orientation proves challenging. Incorporating the 3D Gaussian representation results in a 7\% improvement, demonstrating the feasibility of Gaussian splatting. Further applying 3D Gaussian sampling results in a 1.43\% improvement, confirming that 3D Gaussian sampling captures spatial information more effectively than relying solely on $\texttt{GauAlign}()$.

\visgau
\individualplus
\adaptivecontrol

\textbf{Self-attention and dynamic convolution.} We check the effect of self-attention and dynamic convolution in the decoder, as shown in Table~\ref{table:individaual_attention}. Our method achieves an AP$_{50}$ of 61.91\% without self-attention and dynamic convolution. Sequentially incorporating self-attention results in AP$_{50}$ of 66.21\%. When both of them are applied, the model achieves an AP$_{50}$ of 75.44\%. Additionally, we compare an alternative implementation, replacing dynamic convolution with cross-attention, as shown in Table~\ref{table:individual_cross_attn}. The performance degrades with cross-attention. We argue that the input keys and values here are not processed by a standard transformer encoder.

\visdiffinput

\textbf{3D feature space.} We compare the effect of the 3D feature space with the feature pyramid $\{P_2, \dots, P_6\}$, as shown in Table~\ref{tab:individual_pyramid}. 
In this ablation study, we represent oriented objects with 2D Gaussians and employ $\texttt{RRoIAlign}()$ to extract features from transformed rectangles in $\{P_2, \dots, P_6\}$. 
The results indicate a 2.23\% drop in AP$_{50}$ when using the pyramid, attributed to the lack of spatial information.

\textbf{Gaussian update strategy.} Table~\ref{tab:adaptive control} illustrates the effect of the update strategy for predicted high- and low-confidence 3D Gaussians in adaptive control. The results indicate a 0.66\% drop in AP$_{50}$ with more iterations, when the output of the current step is directly used as input for the next step. We argue that the output distribution is not learned during training. By gradually applying the two update strategies, our GSDet achieves performance gains with only a slight drop in FPS, which verifies the effectiveness of adaptive control.

\textbf{Qualitative results.} We visualize the Gaussians and predictions in Figure~\ref{fig:vis_gau}. The initial Gaussians vary each time due to randomness. After refinement, the Gaussians focus more effectively on potential object areas. The predictions demonstrate that GSDet accurately detects dense, large scale and aspect ratio objects. In addition, Figure~\ref{fig:vis_diff_input} illustrates various randomly initialized Gaussians and their corresponding predictions. We observe that initial Gaussians positioned near objects contribute to the final predictions, and GSDet effectively handles diverse inputs. Interestingly, relatively large-scale Gaussians are able to predict small-sized objects.


\section{Conclusion}
This paper introduces GSDet, a novel detector that is the first to formulate oriented object detection as Gaussian splatting. 
By leveraging anisotropic 3D Gaussians for object representation within a 3D feature space, GSDet effectively captures spatial information through Gaussian sampling to optimize Gaussians' attributes. 3D Gaussians are projected onto the image plane to form 2D Gaussians, which are then transformed into oriented boxes. Benefiting from the random initialization, the proposed Gaussian-to-box paradigm eliminates the reliance on pre-designed object candidates, allowing for adaptive control and dynamic number of Gaussians during inference. Extensive experiments validate the superior performance of GSDet over mainstream methods.

\section*{Acknowledgments}

This work was supported in part by the National Natural Science Foundation of China under Grant 62272461, Grant 62172417, and Grant 62277046; in part by the Double First-Class Project of China University of Mining and Technology for Independent Innovation and Social Service under Grant 2022ZZCX06; in part by the Six Talent Peaks Project in Jiangsu Province under Grant 2015-DZXX-010 and Grant 2018-XYDXX-044; and Funded by the Graduate Innovation Program of China University of Mining and Technology and the Postgraduate Research \& Practice Innovation Program of Jiangsu Province under Grant 123625313.

\bibliographystyle{named}
\bibliography{ijcai25}

\end{document}